\documentclass[a4paper,twoside]{article}

\usepackage{epsfig}
\usepackage{subcaption}
\usepackage{calc}
\usepackage{amssymb}
\usepackage{amstext}
\usepackage{amsmath}
\usepackage{amsthm}
\usepackage{multicol}
\usepackage{pslatex}
\usepackage{apalike}
\usepackage[bottom]{footmisc}
\usepackage{textcomp}
\usepackage{gensymb}
\usepackage{booktabs}
\usepackage{multirow}
\usepackage{caption}
\usepackage[breaklinks=true,bookmarks=false]{hyperref}
\usepackage{SCITEPRESS}     

\newcommand{\authornote}[1]{\begin{minipage}[t]{6.2in} \centering \fontsize{9}{11}\selectfont \textit{ #1} \end{minipage} \\}

\begin{document}

\title{System for 3D Acquisition and 3D Reconstruction using Structured Light for Sewer Line Inspection}

\author{
\authorname{
Johannes Künzel\sup{*}\sup{1}\orcidAuthor{0000-0002-3561-2758},
Darko Vehar\sup{*}\sup{2},
Rico Nestler\sup{2},
Karl-Heinz Franke\sup{2},
Anna Hilsmann\sup{1}\orcidAuthor{0000-0002-2086-0951},
Peter Eisert\sup{1,3}\orcidAuthor{0000-0001-8378-4805}
}
\affiliation{\sup{1}Fraunhofer Institute for Telecommunications, Heinrich Hertz Institute, HHI, Einsteinufer 37, 10587 Berlin, Germany}
\affiliation{\sup{2}Zentrum für Bild- und Signalverarbeitung, Werner-von-Siemens-Straße 12, 98693 Ilmenau, Germany}
\affiliation{\sup{3}Visual Computing Group, Humboldt University Berlin, Unter den Linden 6, 10099 Berlin, Germany}
\email{\{johannes.kuenzel, anna.hilsmann, peter.eisert\}@hhi.fraunhofer.de,\break \{darko.vehar, rico.nester, karl-heinz.franke\}@zbs-ilmenau.de}
\authornote{\sup{*}Johannes Künzel and Darko Vehar have contributed equally.}
}


\abstract{The assessment of sewer pipe systems is a highly important, but at the same time cumbersome and error-prone task.
We introduce an innovative system based on single-shot structured light modules that facilitates the detection and classification of spatial defects like jutting intrusions, spallings, or misaligned joints.
This system creates highly accurate 3D measurements with sub-millimeter resolution of pipe surfaces and fuses them into a holistic 3D model.
The benefit of such a holistic 3D model is twofold: on the one hand, it facilitates the accurate manual sewer pipe assessment, on the other, it simplifies the detection of defects in downstream automatic systems as it endows the input with highly accurate depth information.
In this work, we provide an extensive overview of the system and give valuable insights into our design choices.}

\onecolumn \maketitle \normalsize \setcounter{footnote}{0} \vfill

\section{\uppercase{Introduction}}
Sewage systems only become visible in the life of most people if they break, nevertheless they are one brittle keystone of modern society. 
Therefore, the thorough assessment of sewage systems is immensely important, especially as some of these pipes are older than one hundred years.
Today, mobile robots equipped with cameras are used to provide insights into sewer pipes and highly trained workers evaluate them manually to identify damaged sections.
Many recent contributions tried to alleviate this cumbersome and error-prone annotation task by applying computer vision methods.

There is an ample amount of literature, with most approaches relying directly on monocular images or video sequences as input (often with fisheye lenses), like \cite{xie_automatic_2019,kunzel_automatic_2018,Hansen2015,Zhang2011}, with a recent survey regarding the detection and classification of defects in \cite{li_vision-based_2022}.
But as the literature shows, many previous approaches struggle with the detection of some spatial defects, like for instance misaligned pipe joints or bent pipes, as they are almost impossible to detect without depth information.
Not to mention the even harder problem of classifying such defects by their severeness, as they are often defined by their depth or spatial extension.
Thus, some works tried to reconstruct the 3D structure from 2D images.
In \cite{denzler_reconstruction_2009,kehtarnavaz_time_2010}, the authors exploited the strict vertical movement of the camera for the reconstruction of sewer shafts from monocular fisheye images, whereas in \cite{kannala_measuring_2008} the reconstruction is based on tracked keypoints found on the structured surface.
The authors of \cite{zhang_improving_2021} extended a basic SLAM approach to leverage cylindrical regularity in sewer pipes, showing promising results.
However, as all the aforementioned algorithms rely on tracked keypoints, the generated 3D models are sparse, and the success is susceptible to the illumination conditions and the structures visible in the scene.
In \cite{bahnsen_3d_2021} the authors assess different alternative 3D sensors to facilitate the inspection of sewer pipes.
They compare passive stereo, active stereo and time-of-flight sensors and identify the last one as most suitable, as it works reliably under all simulated lightning conditions and also in the presence of water.
Surprisingly, the authors exclude systems using structured light up-front, due to their ``low popularity'' and ``the poor performance that is expected for this technology under less controlled conditions'' for the lighting. 
So far, research efforts (\cite{Reiling2014,Wang2013,Dong2021}) towards the usage of structured light for the 3d assessment of pipes mostly analyzed systems with a front-facing camera (mostly an ultra-wide or fisheye lens) and a laser projector placed behind, projecting a radial pattern into the camera image.
These systems feature a rather simple and robust design, but are limited in the  resolution of the imaging.
Also, all forward-facing camera systems share a common disadvantage: Areas closest to the camera appear at the edge of the image where the quality of the geometric-optical mapping is the worst. Consequently, all evaluations based on such imagery are suboptimal compared to systems that capture pipe walls perpendicularly.

The only system to our knowledge with a structured lightning solution with a perpendicular view of the pipe surface was presented in \cite{Alzuhiri2021}.
But opposed to our system approach, it only inspects 120\degree{} of the surrounding 360\degree{} pipe surface and also pairs a projector with a stereo camera setup.
They apply a 4-frame phase shifting method using a static projected pattern and a moving camera.
Such an approach limits a system's transverse velocity and consequently its usability and also relies on a precise inertial measurement unit to perform the 3D reconstruction.
We robustly reconstruct the 3D surface from a single camera image of a projected structured light pattern without additional sensor data at a transverse speed of up to 100 mm/s.
It should be noted that most conventional non-3D sewer inspection systems do not exceed this speed.

In this paper, we will demonstrate how some clever design decisions enable the usage of structured light sensors in the harsh conditions of sewer pipes while exploiting their favourable properties to construct 3D pipe models with a sufficient level of detail for characterization and quantification of damages according to DIN-EN 13508, while being robust against low light and pipe materials of homogeneous appearance. We also explain the robust fusion of the individual acquisitions and how the alternative representation, compromised of a cylindrical base mesh, a texture and a displacement map, facilitates the application of downstream processing task.

\section{\uppercase{Hardware design}}
\label{sec:system_design}
\begin{figure}
\begin{center}
   \includegraphics[width=0.99\linewidth]{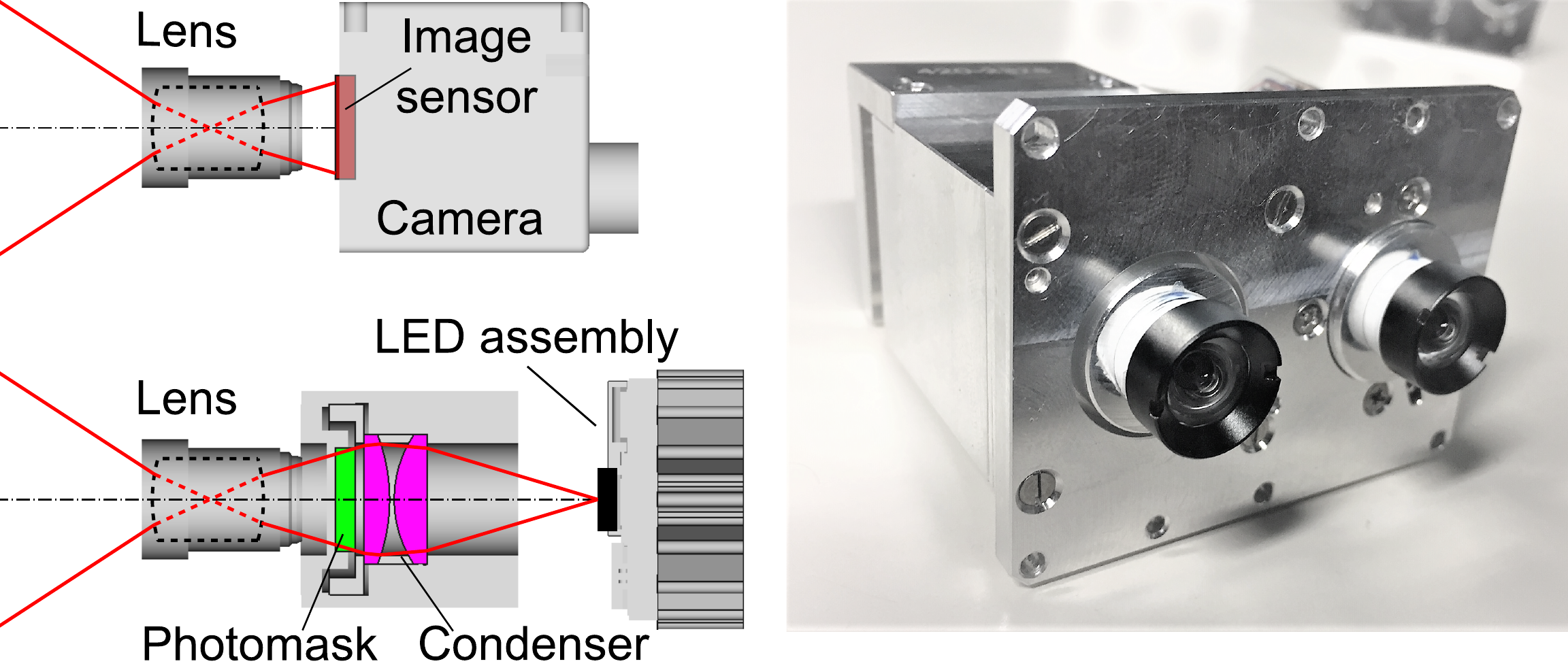}
\end{center}
   \caption{Schematic of the camera projection module (left) and the finished prototype (right) for 3D surface acquisition.}
\label{fig:camera_projection_unit}
\end{figure}

In our setup, each of the six modules (\autoref{fig:camera_projection_unit}) consists of a projection and a camera unit and captures sewer surfaces in 3D according to the principle of structured light. 
The projection and camera units contain the same lenses and have the same parallel offset imaging beam paths. 
The system is designed to capture pipe surfaces with a diameter of 200 mm to 400 mm in 3D with a high-resolution texture using the methods described in \autoref{subsec:single_shot} and \autoref{subsec:texture_processing}. 
Due to the large differences between the minimum and maximum object distances, the field of view (76\degree{} vertical, 61\degree{}  horizontal) and the stereo angle of 29\degree{}, a simulation of optical components was carried out for a careful selection of components to maximize geometric accuracy and image sharpness. 
In particular, the large stereo angle, which results from the size of the hardware components used and usually does not exceed 10\degree{} in comparable applications, poses a challenge for the structured light algorithm on the one hand and leads to a high depth resolution on the other.
The prototype (\autoref{fig:camera_projection_unit} right) captures inner walls of the pipe with a spatial resolution of 0.3 lp/mm and depth resolution of 0.2 mm at $\phi$ 200 mm (1.1 lp/mm and 2.3 mm at $\phi$ 400 mm).

\begin{figure}
\begin{center}
   \includegraphics[width=0.99\linewidth]{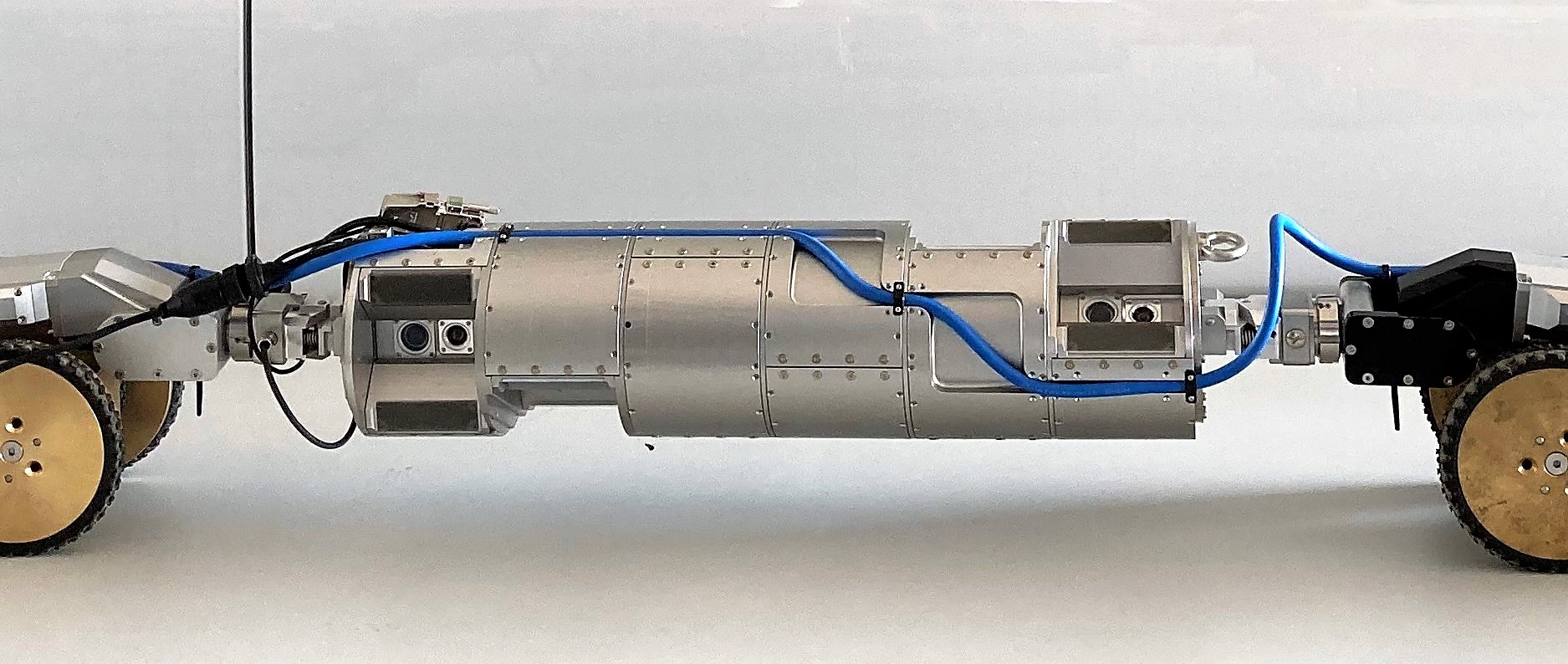}
\end{center}
   \caption{Cylindrical carrier with an outer diameter of 150 mm and length of 600 mm with six structured light modules for 360 degree capture of sewer pipes.} 
\label{fig:funktionsmuster_fahrwagen}
\end{figure}

Six mutually rotated modules are installed in a carrier (\autoref{fig:funktionsmuster_fahrwagen}) with an outer diameter of 150 mm and length of 600 mm for full 360 degree 3D capture of sewer surfaces.
The fields of view of the modules are directed from the central axis to the inner pipe surface and, in their entirety, depict the circumference completely and overlapping, so that adjacent images can be registered and combined.
A vital design decision was the placement of all entrance pupils of the six modules in such a way that they lie on the common carrier axis to ensure this necessary overlapping even with a decentered position and different pipe diameters.

In addition to the pattern projection, we installed a homogeneous texture illumination on each camera module.
Lights for texture and pattern lighting as well as the cameras are hardware controlled and the cameras alternately capture the texture and the projected patterns during continuous travel through the sewer pipe.

\section{\uppercase{3D acquisition}}
\label{sec:methods}

The following section covers the processing chain from raw camera images to the metrically reconstructed 3D point clouds of the inspected sewer section.

\subsection{Single-shot structured light}
\label{subsec:single_shot}

For our single-shot structured light approach, we use binary coded spot patterns based on perfect submaps \cite{Morano1998}. 
In such pseudo-random arrays, each sub-matrix of a fixed size (codeword) occurs exactly once. 
If a pattern based on perfect submaps covers the entire projector image, each projector pixel is uniquely encoded, which is necessary to solve the camera-projector correspondence problem.
We choose a binary instead of a color coding because the projected pattern can be robustly recognized in the camera image without prior knowledge of the color of the captured surfaces.
Additionally, 2D arrays are preferred over 1D sequences when vertical disparity occurs due to non-ideal alignment of the camera and projector planes and in combination with lens distortion.

\begin{figure}[htb]
\begin{center}
   \includegraphics[width=0.99\linewidth]{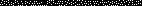}
\end{center}
   \caption{Binary pattern with uniqueness within a window size of 6$\times$6 which is used in the structured light modules. The pattern is repeated horizontally and vertically to fill the entire projector image.}
\label{fig:pattern_6x6_h3}
\end{figure}

It is not necessary that the pattern spans over the entire projector image.
But its length and height should cover the horizontal and vertical disparity range, which depends on the external camera-projector arrangement, the shortest and farthest distances of the objects to be captured and the lens distortion.
Given these conditions, we generated a pattern \autoref{fig:pattern_6x6_h3} of required length and with a uniqueness within a window size of 6$\times$6 using an exhaustive search similar to \cite{Morano1998}. 
In addition to the required codeword uniqueness, we applied constraints on the distribution of points in the pattern to make the decoding of the captured pattern more robust and algorithmic efficient.
These constraints include a minimum hamming distance of 3, a minimum word weight of 4 and 0-connectivity of all non-zero pixels.

In the next step, the image of the projected pattern is decoded, which means that each recognized codeword is linked with its position in the original projector image.
We used a binarization method \cite{Niblack1985} to detect the structured light pattern reliably, in particular on strongly varying backgrounds, e.g. due to cracks or in areas with highly contrasted textures.
The binary image is then scanned with a predefined window size and each bit-string read within the window is trivially found in the lookup table.

3D points are computed from the corresponding points in respective camera and projector images from the decoding step using the linear triangulation method from \cite[p.~312]{Hartley_book_2003}. 
The intrinsic and extrinsic parameters are obtained from the calibration \autoref{subsec:camera_calibration} and the lens distortion is corrected beforehand in normalized image coordinates. 
The resulting 3D point cloud is output as RGB-D image for further processing.


\subsection{Color texture preprocessing and 3D mapping}
\label{subsec:texture_processing}
The spatially resolved detection of materials, visually mapped as color (texture) images, enables the identification of significant regions and also defect features in sewer lines.
In particular, chromaticity is a key feature for detecting and distinguishing defects in downstream processes.
Images in our case are disturbed by color-shading, which represents both chromaticity and brightness shifts. 
For this purpose, it is necessary to correct systematic influences on the color textures caused by lighting, lens, sensor, or variable (unknown) geometry in the primary image data.

\begin{figure}
\begin{center}
   \includegraphics[width=0.99\linewidth]{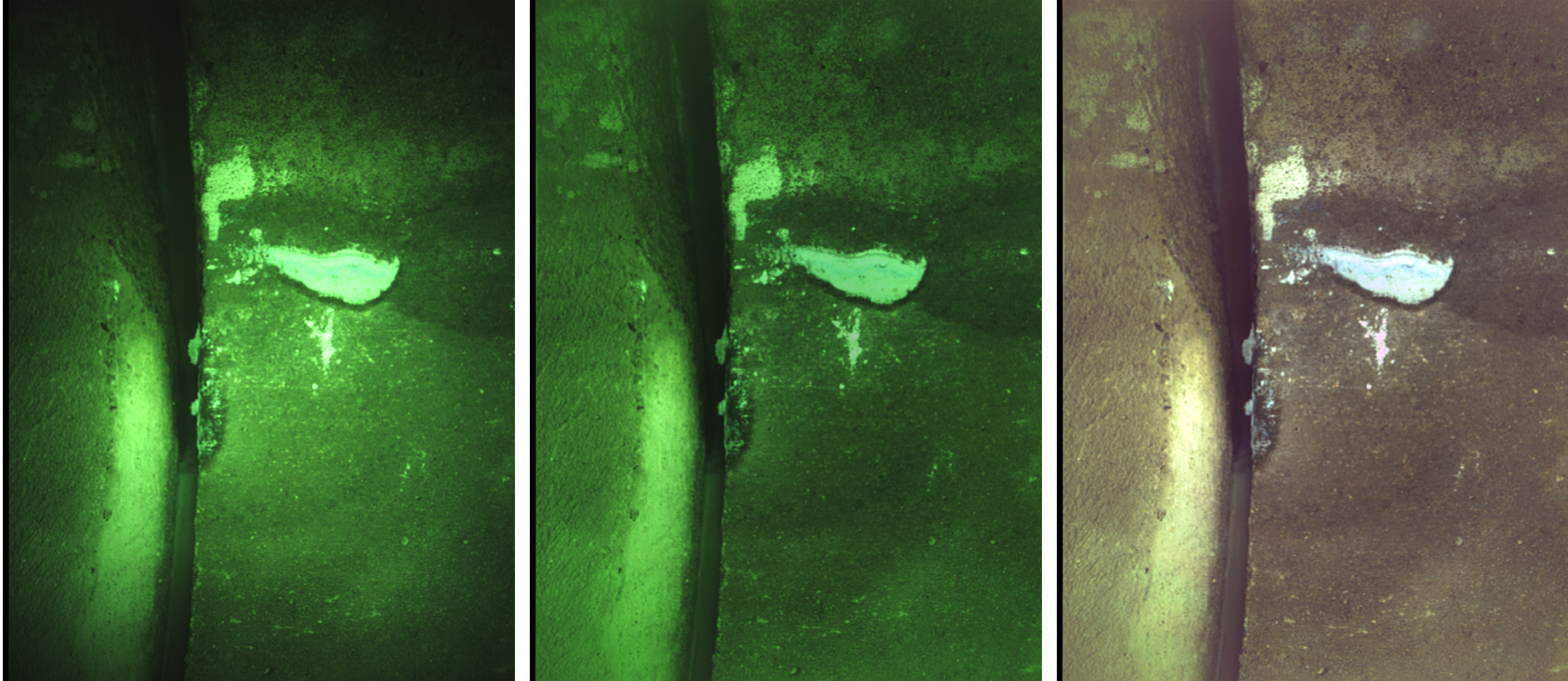}
\end{center}
   \caption{Debayered color image with visible vignetting toward the periphery of the image (left). The result of brightness shading correction (middle) and chromaticity correction (right).}
\label{fig:shading_color_correction}
\end{figure}

Conventional trivial shading correction of the color image by a planar white reference is not applicable as the variable distance between camera and scene objects and the complex shading influences lead to artifacts like chromatic flares.
In order to handle this, we estimate a reference light from images of the captured sewer section and correct color shading with lightness and chromaticity features from L*a*b* color space in all primary color channels. 
The results of this correction can be seen in  \autoref{fig:shading_color_correction}. 

In addition to the color correction we remove motion blur caused by movement during image capture.
By using the prior knowledge of a cylindrical pipe geometry, a depth-graded unfolding using a parametric 1D - Wiener filter \cite{Murli1999} was used to enhance the sharpness of the color texture.

\begin{figure}
\begin{center}
   \includegraphics[width=0.99\linewidth]{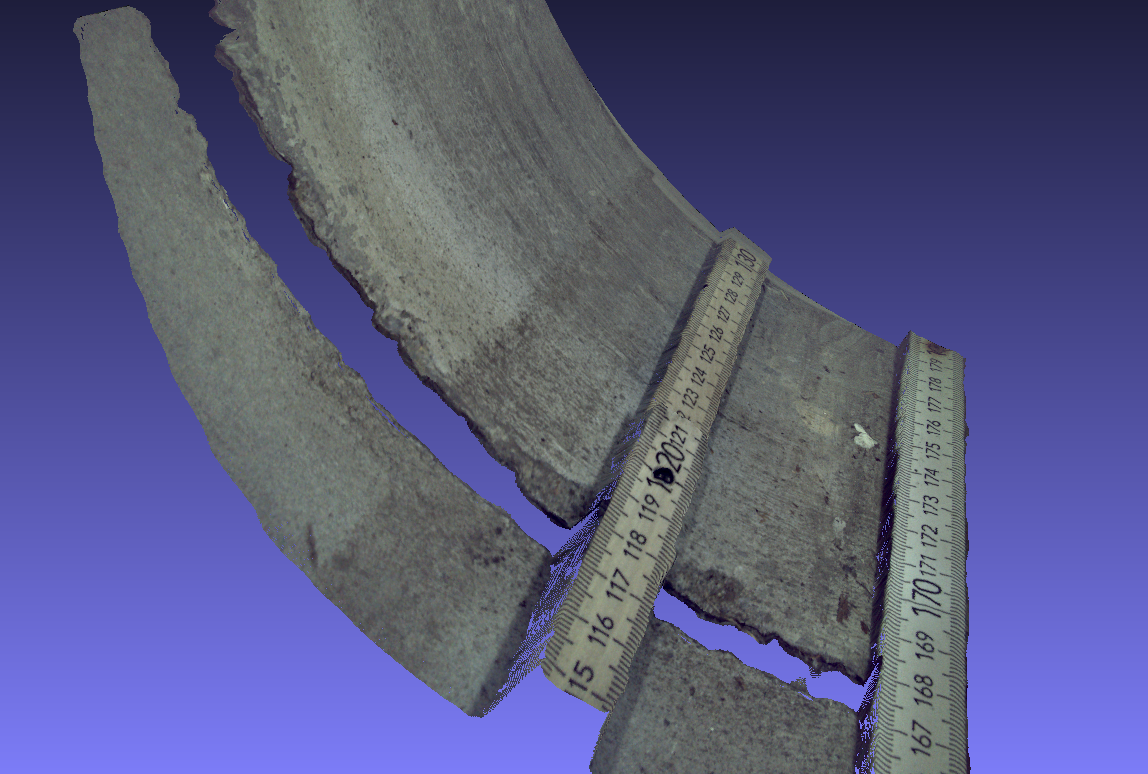}
   
   \vspace{0.5em}
   
   \includegraphics[width=0.99\linewidth]{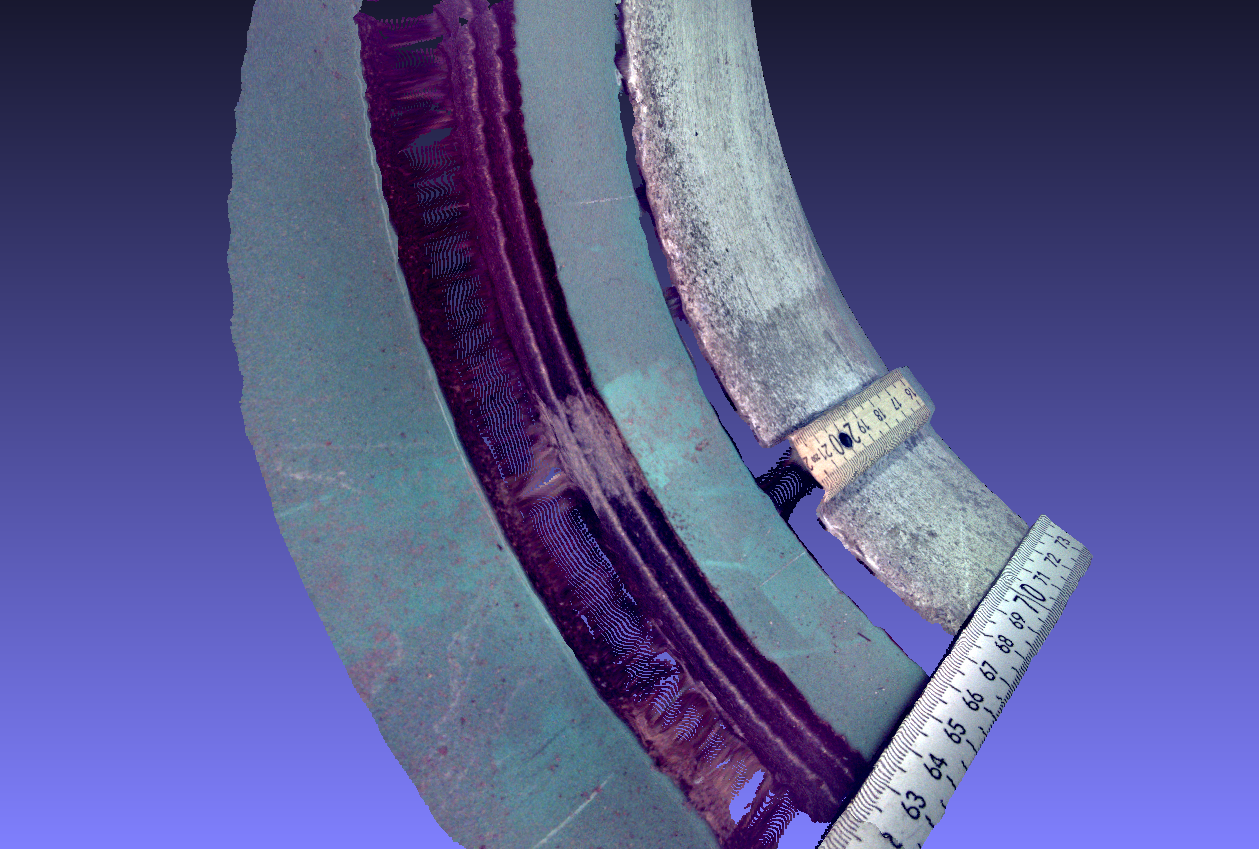}
\end{center}
   \caption{3D color textured point cloud of a connection between two concrete pipes (top) and  between a plastic and concrete pipe (bottom) as captured by a  structured light module. The level of detail in both texture and 3D can be seen on the ruler and tape measure, as well as the pipes stacked at varying depths.}
\label{fig:DN300_concrete_plastic}
\end{figure}

In order to map the preprocessed texture onto the generated 3D point cloud, texture images are captured a few milliseconds before and after the structured light pattern.
Using the camera's intrinsic parameters, sparse feature detection and matching (AKAZE \cite{AKAZE_BMVC2013}) on the texture images and the precise timing of the three-image sequence (texture, 3D, texture), the change in camera pose can be estimated at the time the projected pattern (3D) was captured. 
The resulting transformation is used to map the color texture to the 3D point cloud. 
Results of the proposed mapping procedure for multimodally captured concrete and plastic pipes are shown in \autoref{fig:DN300_concrete_plastic}.

\subsection{Geometric calibration of camera-projector modules}
\label{subsec:camera_calibration}
We approximate the imaging properties of both the camera and the projector (inverse camera) of each structured light module with a pinhole model. 
It is represented by a camera calibration matrix consisting of the effective focal length and the principal point, which is also used as the center of lens distortion. 
To model the significantly observable lens distortion, we use the radial component of a three-term Brown-Conrady distortion model \cite{Brown1966}.

\begin{figure}
\begin{center}
   \includegraphics[width=0.99\linewidth]{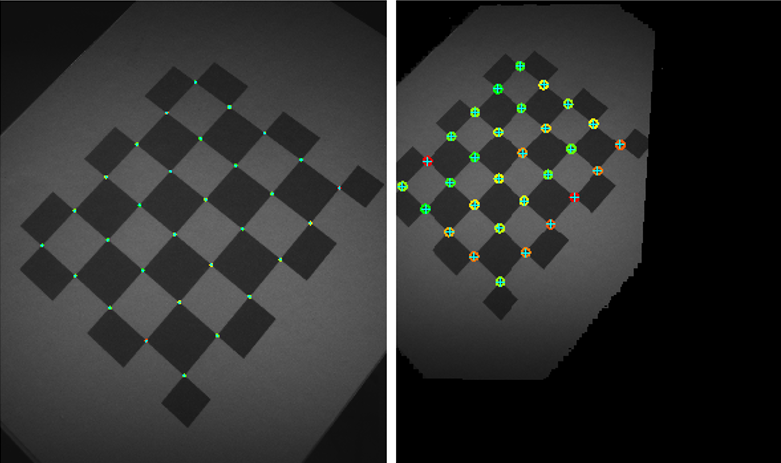}
\end{center}
       \caption{The camera-projector modules are intrinsically and extrinsically calibrated using several images of a calibration pattern taken at different distances and orientations. Detected calibration points in camera (left) and projector (right) are color-coded according to the re-projection error to visually indicate the precision of a calibration.}
\label{fig:camera_projector_calibration_setup}
\end{figure}

We captured 15-20 calibration images for the intrinsic and extrinsic calibration of camera and projector using a tripod-mounted target with a checkerboard pattern as seen in \autoref{fig:camera_projector_calibration_setup}. 
Target detection, sub-pixel processing of calibration points and parameter optimization were carried out with the toolkit 3D-EasyCalib \cite{Vehar2019}.

For intrinsic calibration, the camera matrix is determined from images of a calibration target, its known world points and the corresponding image points.
For both the projector and the extrinsic calibration of a camera-projector module an extra step is needed. 
Instead of a single image, we captured a pair of images for each target pose -- an image of the calibration target with the projected pattern off and the ambient light on and vice versa.
The processing of the target images is done in the same way as for the camera calibration. 
Decoding the pattern gives us the geometric mapping between the image planes of the projector and the camera.
We use local homographies around each checkerboard corner to transfer calibration points between camera and projector images \cite{Moreno2012}.
An identical optimization as for the intrinsic camera calibration was used to estimate the projector calibration matrix as well as the orientation and pose of the projector with respect to the camera (extrinsics).


\label{sec:3dmodel}
\section{\uppercase{Generation of 3D models}}

The 3D measurement process delivers its data in a texture and a depth map  accompanied by the intrinsic camera calibration matrix and the global pose estimated by the odometry ($\mathbf{T}_{i,j}$).
The parameters $i$ and $j$ are the module identifier and the counter of the acquisitions done by this module, respectively. 
The overall goal is to pass on the 3D data to a downstream process for an automatic annotation of defects.
Therefore, we decided to represent our 3D model by a cylindrical mesh, a high-resolution texture, and a displacement map encoding the difference from an ideal cylinder.
This combination is easily digestible to subsequent AI algorithms for defect detection, but can also be rendered for manual assessment.

In the following section, we describe the processing pipeline to generate the output representation from the high-resolution 3D measurements.

%

\subsection{Local registration}

\begin{figure}[htb]
    \centering
    \includegraphics[width=0.8\linewidth]{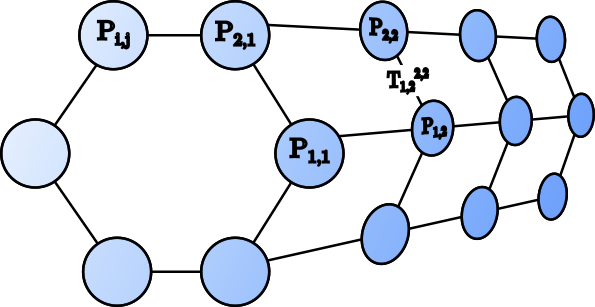}
    \caption{Schematic of the graph-like structure used as foundation of the registration. Each node represents the location of a pipe segment, initialized using the odometry information. The edges represent the pairwise registration between two nodes.}
    \label{fig:graph_mesh}
\end{figure}

In order to register all depth maps, i.e.~sewer pipe segments into a mutual coordinate system, the first step is to calculate relative 3D rigid transformations between neighboring segments, followed by a global optimization into one coordinate systems (next section).
In order to automatically determine the registration targets for the local registrations, we utilize the information about the coarse position of the pipe segment to determine the nearest neighbors.
For each of these, we calculate a mockup point cloud, calculated from the known pipe diameter and the approximate distance between the acquisition module and the pipe surface and project it onto the target point cloud to get a quick estimate of the overlap.
This omits the time consuming creation of the true point cloud from the depth map.
By setting a threshold to separate the neighboring segments, we build the foundation to setup a registration graph like in a Graph-SLAM \cite{grisetti_tutorial_2010} setting, as depicted in \autoref{fig:graph_mesh}, in order to yield consistent registrations between all segments.

With the utilization of the Graph-SLAM framework, we can interpret the global position of each pipe segment as a node in a graph.
The rigid 3D transformation between two of these segments forms a connecting edge.
In the Graph-SLAM literature, this is referred to as a measurement.
We follow two different approaches with mutual benefits in order to relatively register  two individual 3D pipe segments, as described in the following.

\paragraph{Feature-based registration }

As the depth map assigns depth measurement to (almost) each pixel in the texture map, we utilize a sophisticated detection and matching of SIFT features, based on the work of \cite{furch_iterative_2013}.
For the matched feature of the texture map, we extract the associated depth measurements from the depth map.
Pairs with a missing measurement in one or both acquisitions get rejected from the registration.
Subsequently, we estimate the rigid transformation between both point clouds using Singular Value Decomposition.

\paragraph{Projection-based registration}

The feature-based method struggles in regions with low textural and geometrical variation, especially for the case of pipes made of plastic or similar materials, leading to slight misalignments creating unpleasant visual artifacts.
To avoid these, we instead utilize again the projection of the individual pipe segments onto each other to establish 3D correspondences in order to calculate the transformation between them.
This method heavily relies  on the information from the odometry, which defines the initial placement of the segments and therefore their overlap.
The odometry in our scenario, on the other hand, has access to a sensor using a cable between the robot and the base station to measure the traveled distance and makes the described registration therefore an favorable alternative in the absence of distinctive visual features.

\subsection{Global non-linear quadratic optimization}

To optimize the global registration of the segments, we formulate our problem as a pose graph-based SLAM problem in three dimensions.
The initial poses of the different segments are given by the odometry as
$$
\mathbf{P}_{i,j} = [ \mathbf{p}_{i,j}^T, \mathbf{q}_{i,j}^T ]^T,
$$
where $\mathbf{p}$ is a 3D vector representing the position and $\mathbf{q}$ is a quaternion representing the orientation.
The relative pose constraints delivered by the pairwise relative registrations are given by
$$
\mathbf{T}_{i,j}^{k,l} = [(\mathbf{p}_{i,j}^{k,l})^T, (\mathbf{q}_{i,j}^{k,l})^T]^T = [\hat{\mathbf{p}}^T, \hat{\mathbf{q}}^T]^T,
$$
where $k$ and $l$ denote the module and segment identifier of the query segment, respectively, and $\hat{\mathbf{p}}$ and $ \hat{\mathbf{q}}$ are used as a concise representation.
Please refer to \autoref{fig:graph_mesh} for a visualization of the graph for the global registration.
The residual can then be calculated by
$$
\delta =
\begin{bmatrix}
R(\mathbf{q}_{i,j})^T(\mathbf{p}_{k,l}- \mathbf{p}_{i,j})-\hat{\mathbf{p}} \\
2vec((\mathbf{q}_{i,j}^{-1}\mathbf{q}_{k,l})\hat{\mathbf{q}}^{-1})
\end{bmatrix}
$$
with $\hat{\mathbf{p}}$ and $\hat{\mathbf{q}}$ being a concise notation of the relative pose constraint between the two involved segments, $vec(\mathbf{q})$ the vector part and $R(\mathbf{q})$ the rotation matrix of the quaternion.
Additionally, the residuals can be weighted by the covariance matrices of the registrations.
Thus, we can take into account the lower reliability of the projection-based registration.
We also incorporate prior knowledge into the graph pose estimation, as we expect the pipes to be approximately straight.
To do so, we add additional constraints between the first and the last acquisition of each module.
We solve the equation system with Ceres \cite{Agarwal_Ceres_Solver_2022} to get the final poses of the different segments $\mathbf{T}_{i,j} = [ \mathbf{p}_{i,j}^T, \mathbf{q}_{i,j}^T ]^T$.

\subsection{Integration of uncertainty and prior knowledge}

The Graph-SLAM framework \cite{grisetti_tutorial_2010} enables us to associate each measurement with an estimation of the uncertainty, by weighting the pose parameters with a covariance matrix.
Thus, we weight the feature-based registrations higher than the projection-based, accounting the expected higher precision.
We also use this re-weighting, the odometry information and the assumption of straight pipes to introduce additional regularization edges between the first and the last segment of each module.
To this end, we manually set the entries in the respective covariance matrices to constrain the rotation and translation.

\subsection{Creation of texture and displacement maps}

Our final representation comprises a cylindrical mesh with a detailed texture and displacement map, facilitating e.g.,the following AI algorithms, as they can easily analyze these maps.
To this end, we fit a cylinder to the registered point clouds of the individual pipe segments.
Afterwards, we transform all points accordingly to align the model of the pipe with one axis of the Euclidean coordinate system and transform the 3D coordinates of the points into cylindrical coordinates, so that each point $\mathbf{P}$ of the point cloud $\mathbf{C}_{i,j}$ is defined by $P = (\rho, \phi, z)$, with $\rho$ representing the axial distance, $\phi$ the azimuth and $z$ the axial coordinate.
Depending on the desired output resolution, we define an equally spaced grid, with the size of each grid cell being set to the desired resolution (for example 0.1mm) and utilizing the cylindrical coordinates, assign each point to one grid cell.
Depending on the resolution of the input point clouds, and the desired output resolution, the number of points assigned to each cell varies.
The final axial distance and texture gets calculated by the mean over all points assigned to the respective cell.
With a spatial resolution of 0.2 mm, the resulting texture and displacement maps for 0.5m of modelled pipe have a resolution of 2500 pixels (dimension along the pipe axis) and 4965 pixels (along the circumference of the pipe). 

\section{\uppercase{Results}}

\subsection{Evaluation of the 3D acquisition}
\label{subsec:eval_acq}
In order to demonstrate the consistency and reliability of the 3D measurement for a module, we captured and evaluated ruled and real geometries (planes and cylinders) based on the distance of a plane to the center of the camera, the radius of the cylinder and the overall error of the 3D reconstruction compared to the ideal model.

In particular, we analyze the accuracy of the determined intrinsic and extrinsic parameters of the modules as well as the consistency and suitability of the 3D measurements to support the subsequent defect detection.

\subsubsection{Experimental setup}

For the plane fitting, an ideally white and planar board was positioned in front of the module and perpendicular to the principal axis of the camera (projector). 
The board was captured at the approximate distances of 90 mm and 140 mm to the camera center, representing flat sections of the used 3D measuring space. 
Furthermore, three pipes with diameters of 200 mm, 300 mm, and 400 mm were captured in 3D.
The modules were manually placed in the same positions as possible and aligned visually. 
The exact definition of the object distance from the view of the later application cases of the module is irrelevant.
Contrary to the measured planes, the vitrified clay and concrete pipes deviate slightly from ideal cylinders and have artificially added point- and line-shape defects with a depth of 1-2 mm. 
We determine the parameters of the plane and cylinder using the least squares method from a non-defect-free subset (10$\times$10 grid) of the reconstructed 3D point clouds. 
We evaluate the Euclidean distances of the measured points (integral shape deviations characterized by RMS and maximum errors) from the ideal plane or the ideal cylinder. 
To check the plausibility, we additionally determine distances of the camera to the plane and cylinder radii.

\begin{figure*}[t]
     \centering
     \begin{subfigure}[t]{0.48\textwidth}
         \centering
         \includegraphics[width=\textwidth]{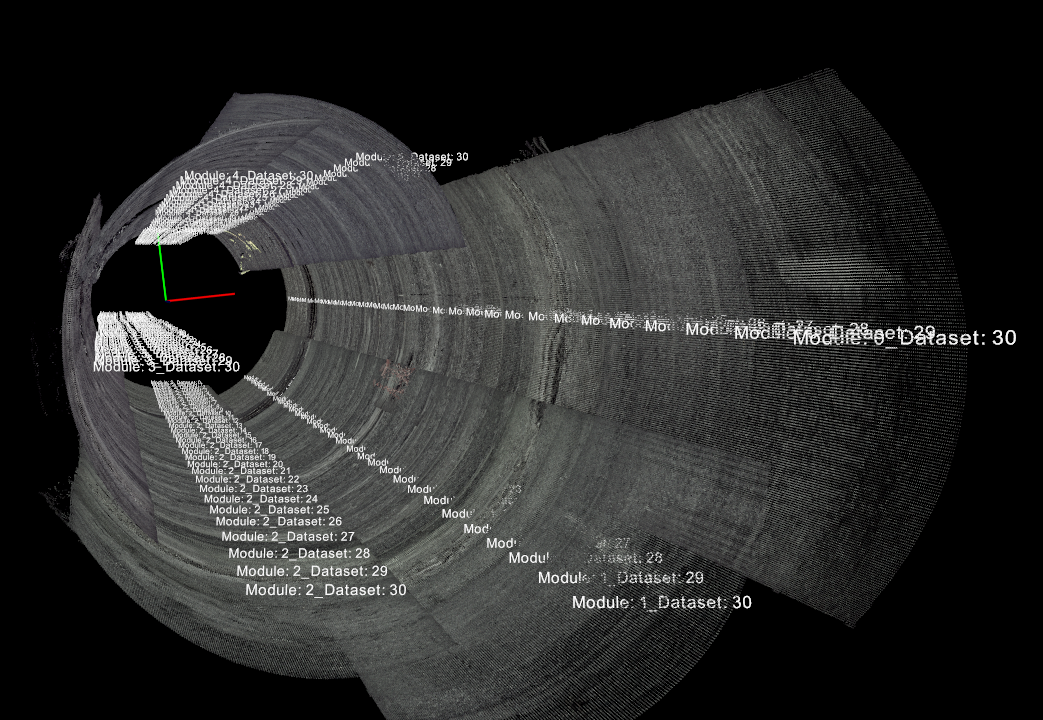}
         \caption{Initial placement.}
         \label{fig:pre_registration}
     \end{subfigure}
     \hfill
     \begin{subfigure}[t]{0.48\textwidth}
         \centering
         \includegraphics[width=\textwidth]{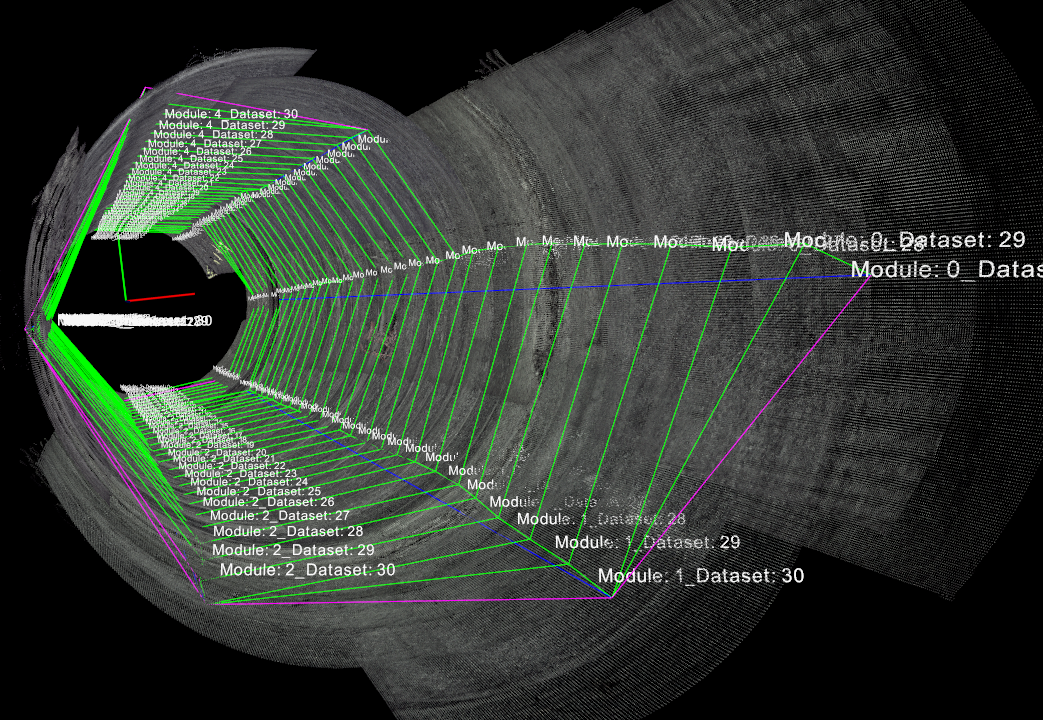}
         \caption{After registration.}
         \label{fig:post_registration}
     \end{subfigure}
        \caption{Sample result of the registration of 30 consecutive pipe segments. The initial configuration is depicted on the left side, with strongly visible misalignments, especially around the circumference. The right side shows the final configuration after the registration (green edges: feature-based, purple edges: projection-based, blue edges: regularization edges).}
        \label{fig:results_registration}
\end{figure*}

\subsubsection{Results}

\begin{figure}
\begin{center}
   \includegraphics[width=0.99\linewidth]{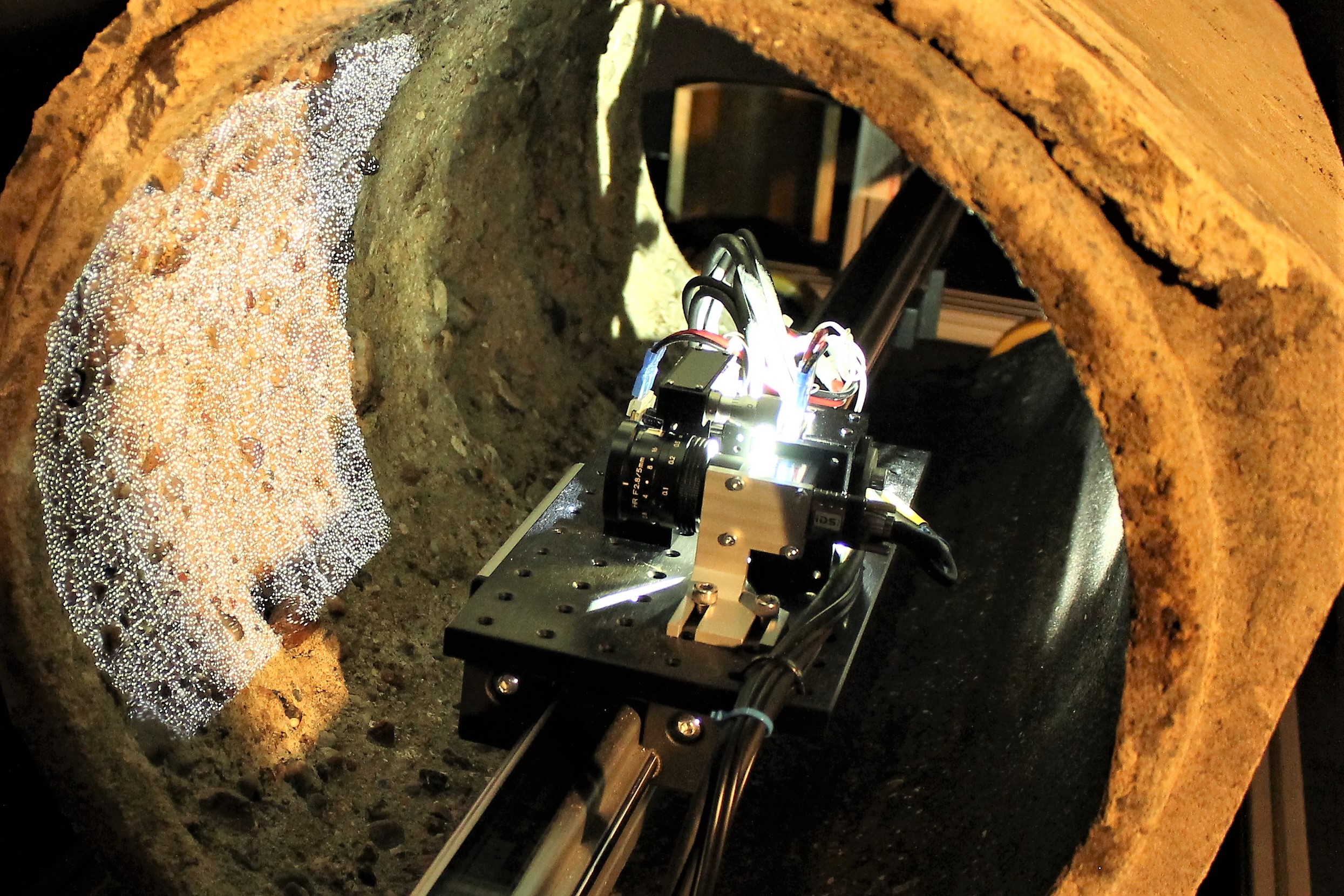}
\end{center}
   \caption{Experimental setup for evaluating the generated 3D point cloud of a real sewer section. Please note that the rig shown was only used to evaluate different camera-projector setups during our research. The final robot is depicted in \autoref{fig:funktionsmuster_fahrwagen}. }
\label{fig:plane_cylinder_evaluation}
\end{figure}

\begin{table}
\caption{Median distance, RMS and maximum error of the captured point clouds to the fitted plane for each camera-projector module. Distances, RMS and maximum errors are given in millimeters.}
  \begin{center}
    {\small{
\begin{tabular}{lcccc}
\toprule
 & Module \# & median(d) & RMS err. & Max err. \\
\midrule
\multirow{6}{*}{\rotatebox[origin=c]{90}{d = 90 mm}}& 1 & 90.91 & 0.09 & 0.28\\
& 2 & 89.70 & 0.10 & 0.30\\
& 3 & 90.32 & 0.07 & 0.36\\
& 4 & 89.94 & 0.07 & 0.64\\
& 5 & 89.89 & 0.06 & 0.29\\
& 6 & 89.81 & 0.07 & 0.45\\
\midrule
\multirow{6}{*}{\rotatebox[origin=c]{90}{d = 140 mm}}& 1 & 140.54 & 0.12 & 0.58\\
& 2 & 139.25 & 0.15 & 0.58\\
& 3 & 140.42 & 0.16 & 0.59\\
& 4 & 140.08 & 0.14 & 0.62\\
& 5 & 140.19 & 0.14 & 0.71\\
& 6 & 139.58 & 0.17 & 0.71\\
\bottomrule
\end{tabular}
}}
\end{center}
\label{tab:plane_evaluation}
\end{table}

\begin{table}
\caption{Estimated cylinder diameter from captured point clouds by each camera-projector module. The errors in the last two columns refer to the deviation of all 3D points from the fitted cylinder. Distances, RMS and maximum errors are given in millimeters.}
  \begin{center}
    {\small{
\begin{tabular}{lcccc}
\toprule
 & Module \# & Diameter & RMS err. & Max err.\\
\midrule
\multirow{6}{*}{\rotatebox[origin=c]{90}{$\phi$ 200 mm}}& 1 & 200.50 & 0.09 & 0.68\\
& 2 & 201.04 & 0.10 & 0.94\\
& 3 & 200.96 & 0.08 & 0.83\\
& 4 & 202.32 & 0.09 & 1.00\\
& 5 & 201.95 & 0.09 & 0.47\\
& 6 & 201.98 & 0.10 & 0.57\\
\midrule
\multirow{6}{*}{\rotatebox[origin=c]{90}{$\phi$ 300 mm}}& 1 & 300.58 & 0.18 & 0.90\\
& 2 & 298.58 & 0.19 & 1.01\\
& 3 & 299.73 & 0.18 & 0.99\\
& 4 & 302.09 & 0.18 & 0.93\\
& 5 & 302.73 & 0.17 & 1.08\\
& 6 & 301.10 & 0.17 & 0.97\\
\midrule
\multirow{6}{*}{\rotatebox[origin=c]{90}{$\phi$ 400 mm}}& 1 & 402.34 & 0.30 & 2.07\\
& 2 & 394.96 & 0.28 & 2.42\\
& 3 & 396.38 & 0.26 & 1.98\\
& 4 & 403.76 & 0.32 & 2.83\\
& 5 & 401.99 & 0.29 & 1.98\\
& 6 & 397.42 & 0.29 & 2.20\\
\bottomrule
\end{tabular}
}}
\end{center}
\label{tab:cylinder_evaluation}
\end{table}

\begin{figure}
\begin{center}
   \includegraphics[width=0.55\linewidth]{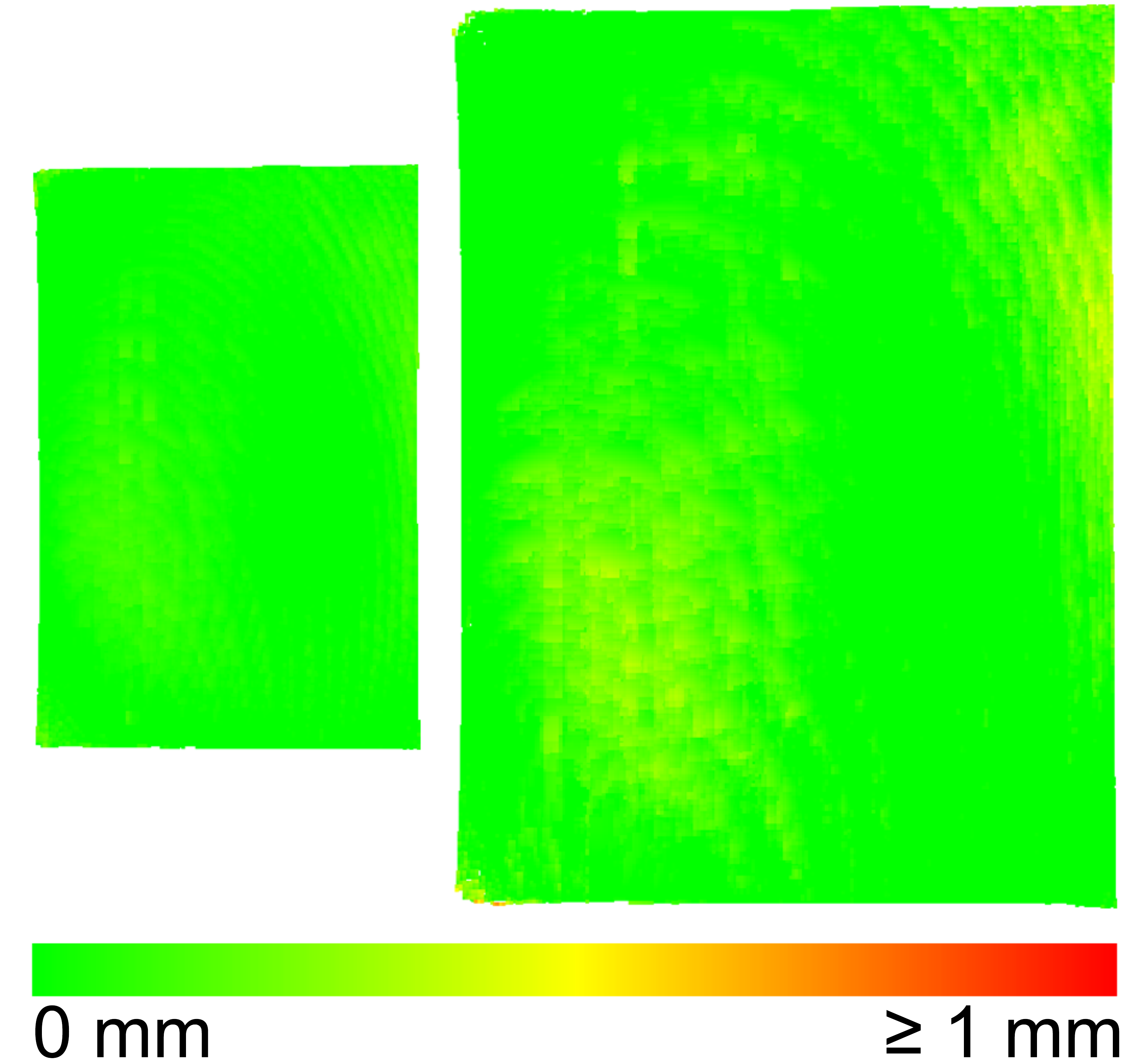}
\end{center}
   \caption{Color coded difference of 3D points to an ideal plane at 90 mm (left) and 140 mm (right) distance. The deviations visible as yellow areas on the right are caused by the imprecise radial distortion parameters.}
\label{fig:plane_90_140}
\end{figure}

\begin{figure}
\begin{center}
   \includegraphics[width=0.99\linewidth]{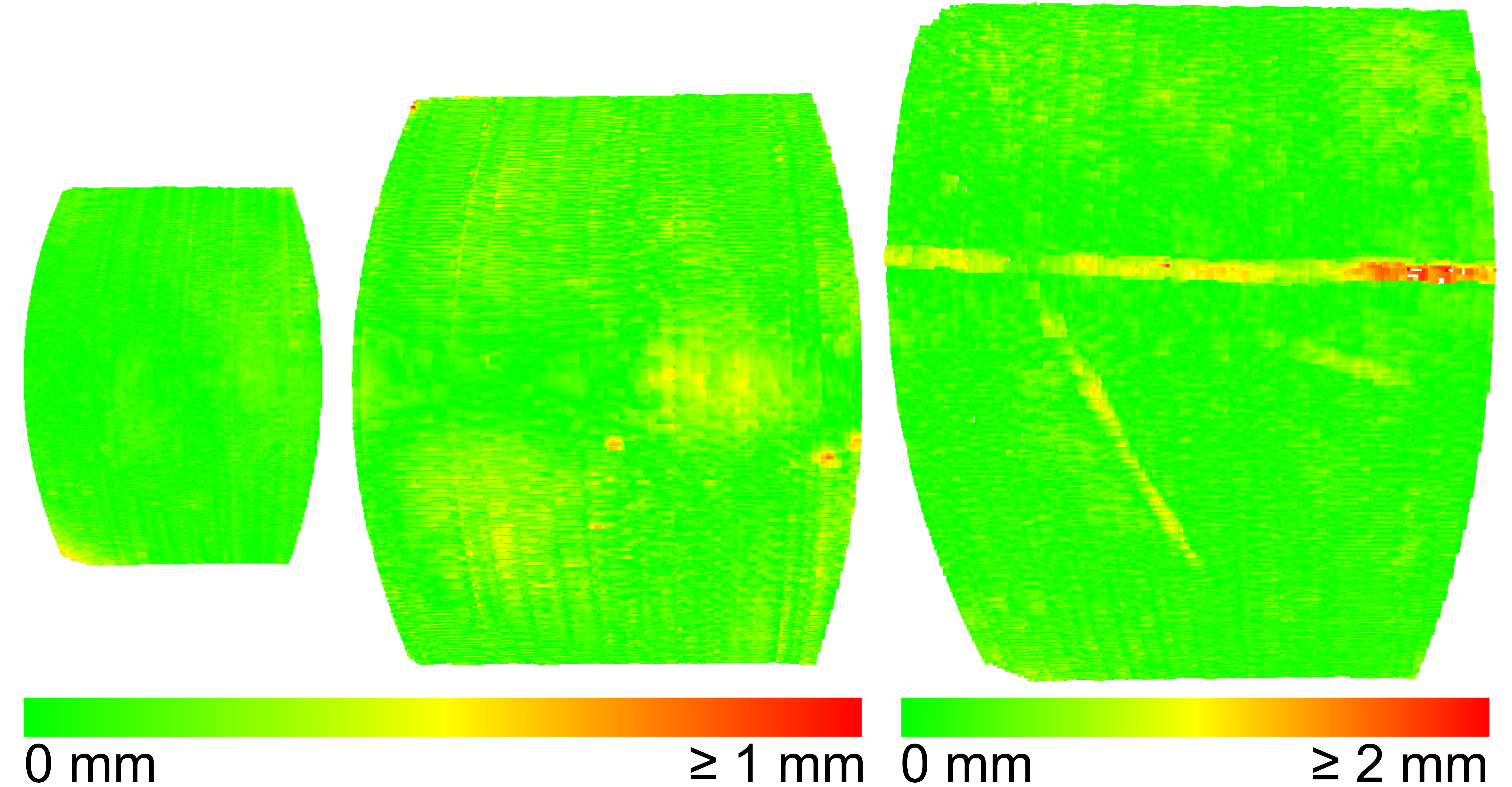}
\end{center}
   \caption{Color coded difference of 3D points to a fitted 200 mm (left), 300 mm (middle) and 400 mm (right) diameter cylinder. Both real shape defects and surface damages can be seen clearly. Note the different color coding of the largest cylinder on the right.}
\label{fig:cylinder_200_400.pdf}
\end{figure}

The resulting medians of the distances in \autoref{tab:plane_evaluation} correspond to the real distances of the manually aligned and positioned plane. 
Very small RMS errors are due to the large stereo angle and precise system calibration.
The maximum error of the plane (d = 90 mm) for each module was observed at the corner of the image.
At a distance d = 140 mm, surface deviations can be seen as yellow areas in \autoref{fig:plane_90_140}  (right) which occur in a similar way with other modules.
These two observations are the result of an imprecisely determined lens distortion model.

The results of the second experiment, in which pipes of different diameters were measured, are shown in \autoref{tab:cylinder_evaluation}.
The estimated diameters correspond very well to the real ones. 
Since only approximately 1/6 of the entire pipe circumference was captured each time, the errors can also be attributed to the cylinder fitting.
The RMS and maximum errors for 200 mm diameter cylinders correspond to the values from the experiment with the plane at 90 mm.
The values for the $\phi$ 300 mm and 400 mm are further skewed by real shape defects and damages to the surface which can be seen in the middle and the right image of \autoref{fig:cylinder_200_400.pdf}.
The image of the 400 mm cylinder clearly shows that the module was not positioned in the center of the tube, but this does not affect the results shown in the table.

We can conclude that the intrinsic and extrinsic parameters have been precisely determined and that the resulting 3D reconstructions of each module meets the accuracy requirement. 
With the help of the data from the 3D module, defects can be described more meaningfully using 3D information and defect areas with geometric properties could be detected more easily and reliably.

\begin{figure*}[htb]
     \centering
     \begin{subfigure}[tb]{0.99\textwidth}
         \centering
         \includegraphics[width=\textwidth]{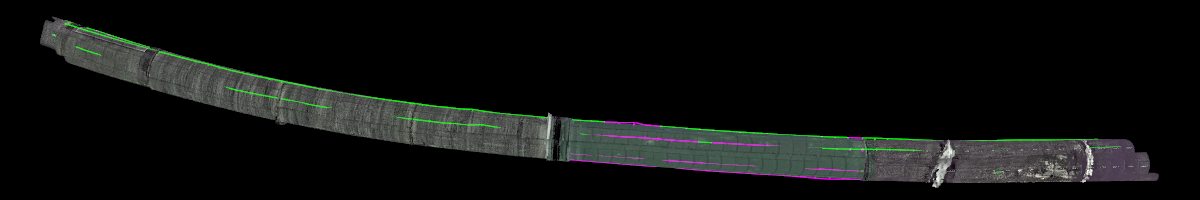}
         \label{fig:wo_regularisation}
     \end{subfigure}
     
     \begin{subfigure}[tb]{0.99\textwidth}
         \centering
         \includegraphics[width=\textwidth]{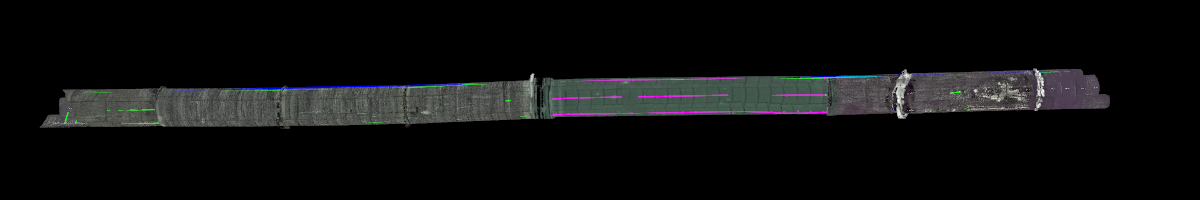}
         \label{fig:w_regularisation}
     \end{subfigure}
        \caption{Example image showing an apparently bent pipe, caused by the accumulation of small local registration errors (upper image) and the corrected geometry, due to the additional regularization edges (blue, lower image).}
        \label{fig:results_regularisation}
\end{figure*}

\begin{figure*}[tb]
     \centering
     \begin{subfigure}[tb]{0.3\textwidth}
         \centering
         \includegraphics[width=\textwidth]{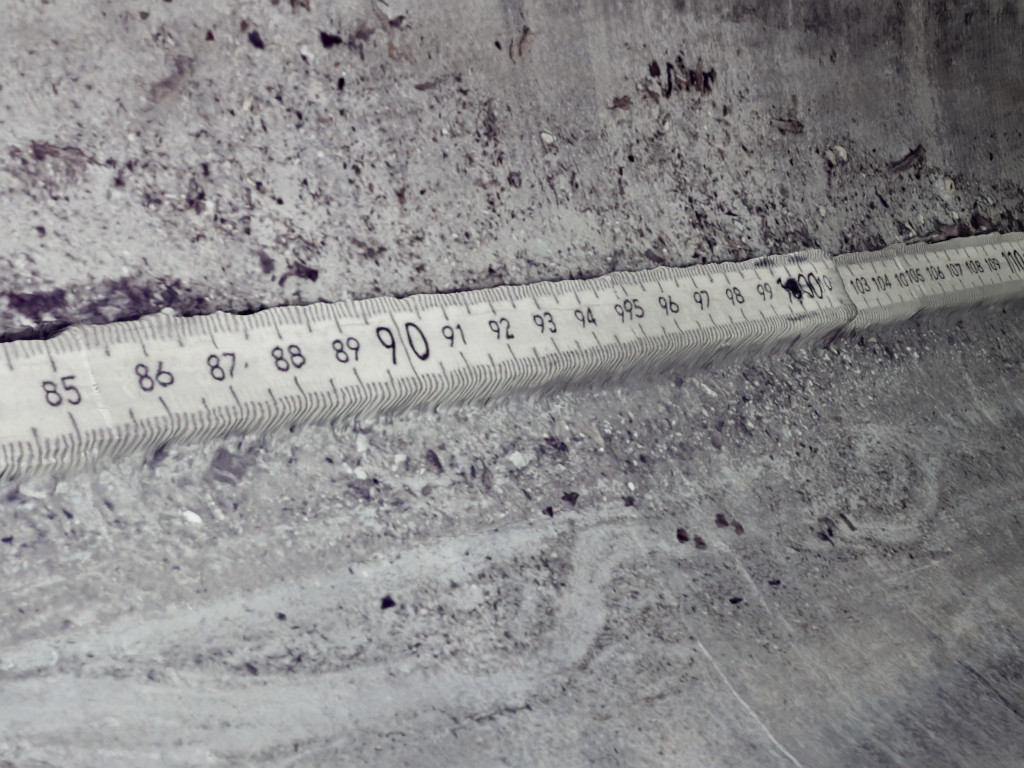}\
     \end{subfigure}
     \hfill
     \begin{subfigure}[tb]{0.3\textwidth}
         \centering
         \includegraphics[width=\textwidth]{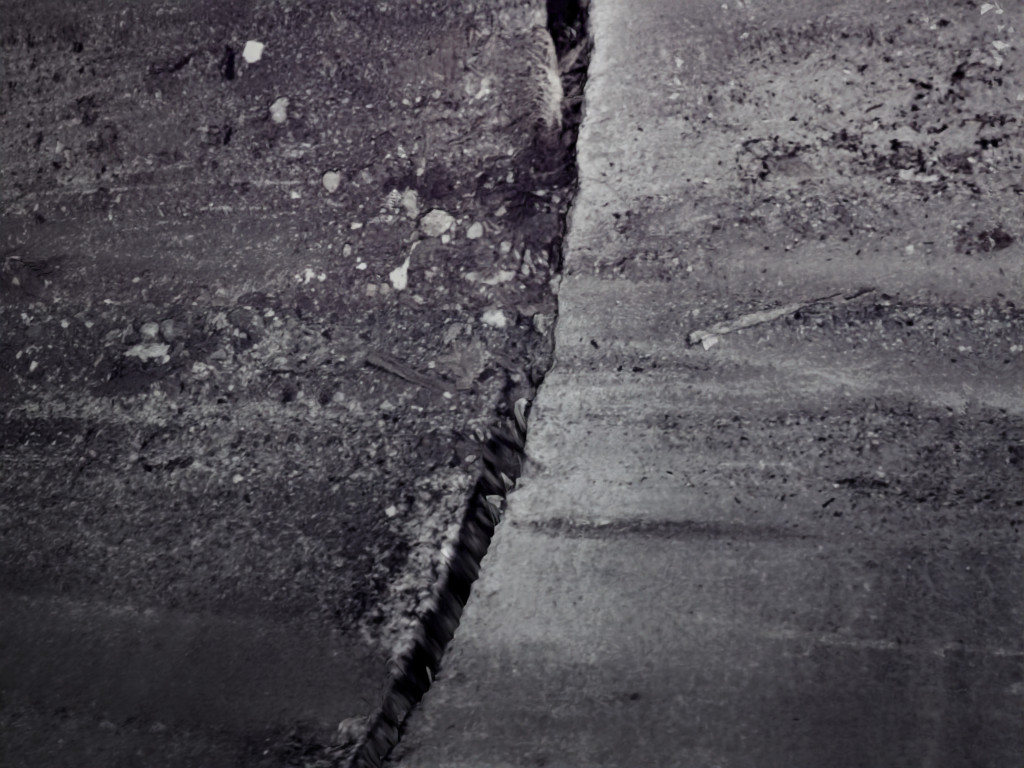}
     \end{subfigure}
     \hfill
     \begin{subfigure}[tb]{0.3\textwidth}
         \centering
         \includegraphics[width=\textwidth]{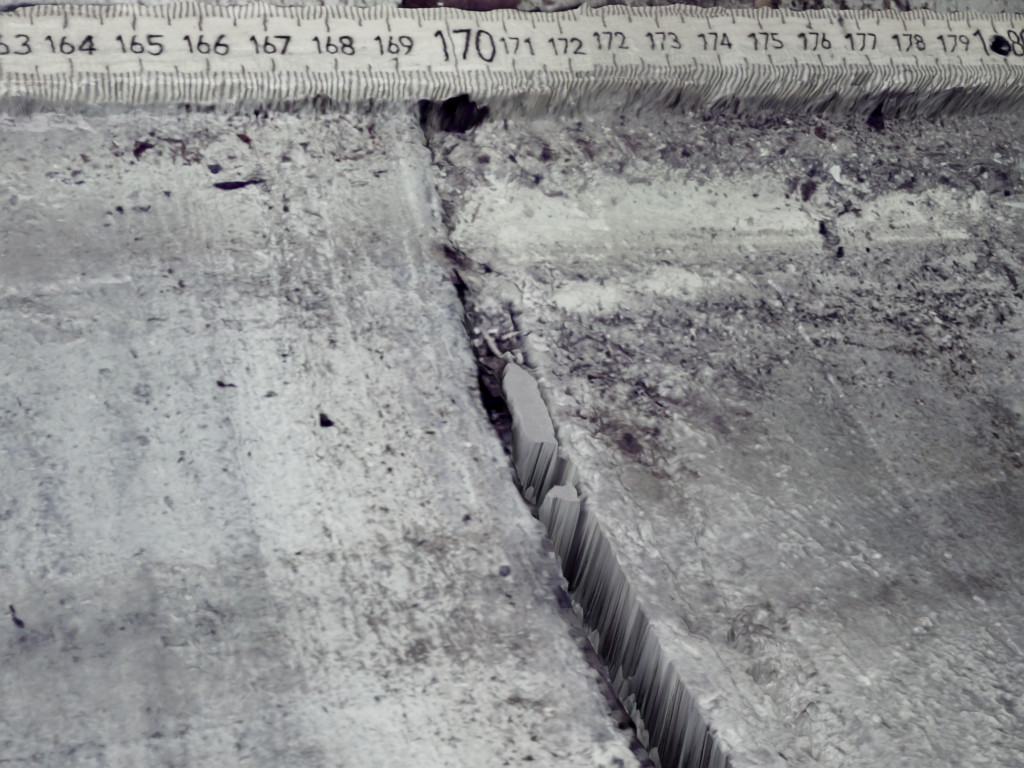}
     \end{subfigure}
     \caption{Samples results rendered using the texture and displacement maps of the individual pipes with subdivision of the initial cylindrical mesh.}
     \label{fig:results_rendering}
\end{figure*}

\subsection{Evaluation of the registration and modelling}
\label{subsec:eval_3d_model}

We acquired test data from several test drives using the developed robot equipped with the proposed single-shot structured light modules.
The measurement of the precise actual 3D geometry of the test pipes is a substantial challenge, which we could not tackle in the scope of this work.
We therefore present and discuss some representative qualitative results.
An example of the registration process can be seen in \autoref{fig:results_registration}.
The unequal length of the different stripes is due to the special arrangement of the modules, as described in \autoref{sec:system_design}.
After the registration, the different segments got aligned successfully, mostly by utilizing the feature-based registration.
Only at the end, the projection-based registration method came into operation, as the overlap was too small to identify and match enough feature points.

The images in \autoref{fig:results_regularisation} showcase the influence of the regularization edges.
The visible bending is mainly caused by the plastic pipe section, which exhibits almost no visual or spatial features to track.
Therefore, the global registration accumulates errors from the projection-based registrations utilized there, resulting in a apparently bent pipe.
The regularization edges correct this issue, by constraining the allowed deviation from a straight pipe.

To visualize the final 3D models, we configured a computer graphics suite to apply the texture and displacement map to the cylindrical mesh.
We used subdivision for the coarse mesh triangles to account for the high-resolution displacement maps. Some representative rendered images are depicted in \autoref{fig:results_rendering} highlighting the fine resolution and the metric measurement of the sewer pipe.

\section{\uppercase{Conclusion}}
\label{sec:conclusion}

We presented a sophisticated approach to generate highly accurate metric 3D models of sewer pipes, using specifically designed single-shot 3D modules.
Our work demonstrates how an intelligent design and elaborated algorithms can bring structured light techniques into action, despite the rugged conditions present.
The availability of highly accurate sewer pipe models endowed with precise depth information enables an abundance of new working directions and opens up new possibilities for pattern recognition on this data.
Obviously, they can support and simplify image processing and, in this form, can be easily made available to neural networks for defect detection.
We also strongly believe that they also enable the accurate classification of the detected features, based on their geometric and optical characteristics.
Additionally, they can be also used to train generative processes for the production of artificial training data, to satisfy the demand of data-driven approaches.
We see our further research directions in the system's minimization and the proof of the enhancement of the detection results when depth information is used as an additional input.

\section{Acknowledgement}
This work is supported by the German Federal Ministry of Education and Research (Auzuka, grant no. 13N13891) and by the German Federal Ministry for Economic Affairs and Climate Action (BIMKIT, grant no. 01MK21001H).

\bibliographystyle{apalike}
{\small
\bibliography{biblatex-wacv-auzuka}}

\end{document}